\documentclass{article}

     \usepackage[preprint]{neurips_2019}

\usepackage[utf8]{inputenc} 
\usepackage[T1]{fontenc}    
\usepackage{hyperref}       
\usepackage{url}            
\usepackage{booktabs}       
\usepackage{amsfonts}       
\usepackage{nicefrac}       
\usepackage{microtype}      
\usepackage{amsmath}
\usepackage{mathtools,etoolbox}
\DeclarePairedDelimiterX{\abs}[1]{\lvert}{\rvert}{\ifblank{#1}{{}\cdot{}}{#1}}

\usepackage{caption}
\usepackage{subcaption}

\providecommand{\keywords}[1]{\textbf{\textit{Index terms---}} #1}

\usepackage{natbib}

\title{ Personalised recommendations of sleep behaviour with neural networks using sleep diaries captured in Sleepio\textsuperscript{TM} }
\author{%
  Alejo Nevado-Holgado \\
  University of Oxford \\
  \texttt{alejo.nevado-holgado@psych.ox.ac.uk} \\
  \And
  Colin Espie \\
  Sleepio\textsuperscript{TM} \\
  \texttt{colin@sleepio.com} \\
  \And
  Maria Liakata \\
  University of Warwick \\
  \texttt{m.liakata@warwick.ac.uk} \\
  \And
  Alasdair Henry \\
  Sleepio\textsuperscript{TM} \\
  \texttt{alasdair.henry@sleepio.com} \\
  \And
  Jenny Gu \\
  Sleepio\textsuperscript{TM} \\
  \texttt{jenny.gu@sussex.ac.uk} \\
  \And
  Niall Taylor \\
  University of Oxford \\
  \texttt{niall.taylor@st-hughs.ox.ac.uk} \\
  \And
  Kate Saunders \\
  University of Oxford \\
  \texttt{kate.saunders@psych.ox.ac.uk} \\
  \And
  Tom Walker \\
  Big Health \\
  \texttt{tom.walker@bighealth.com} \\
  \And
  Chris Miller \\
  Sleepio\textsuperscript{TM} \\
  \texttt{chris.miller@bighealth.com} \\
}

\begin{document}

\maketitle
\begin{abstract}
Sleepio\textsuperscript{TM} is a digital mobile phone and web platform that uses techniques from cognitive behavioural therapy (CBT) to improve sleep in people with sleep difficulty. As part of this process, Sleepio captures data about the sleep behaviour of the users that have consented to such data being processed. For neural networks, the scale of the data is an opportunity to train meaningful models translatable to actual clinical practice. In collaboration with Big Health, the therapeutics company that created and utilizes Sleepio, we have analysed data from a random sample of 401,174 sleep diaries and built a neural network to model sleep behaviour and sleep quality of each individual in a personalised manner. We demonstrate that this neural network is more accurate than standard statistical methods in predicting the sleep quality of an individual based on his/her behaviour from the last 10 days. We compare model performance in a wide range of hyperparameter settings representing various scenarios. We further show that the neural network can be used to produce personalised recommendations of what sleep habits users should follow to maximise sleep quality, and show that these recommendations are substantially better than the ones generated by standard methods. We finally show that the neural network can explain the recommendation given to each participant and calculate confidence intervals for each prediction, all of which are essential for clinicians to be able to adopt such a tool in clinical practice.
\end{abstract}


\section{Introduction}
\label{sec:introduction}
Thanks to the widespread use of digital platforms, sleep diaries in the form of smartphone, web and computer applications have become remarkably useful clinical tools to monitor sleep behaviour \cite{buysse_recommendations_2006, carney_buysse_ancoli-israel_edinger_krystal_lichstein_morin_2012}. These platforms have a series of advantages compared to traditional monitoring methods\cite{monk_pittsburgh_1994, haythornthwaite_development_1991}, including: their portability, ease of use, low cost, and the frequency with which they can collect information from the user. These advantages make them an ideal tool to monitor psychological interventions, where they are increasingly being applied \cite{carney_buysse_ancoli-israel_edinger_krystal_lichstein_morin_2012, luik_digital_2017}. 

Sleepio in particular is based on digital CBT interventions for insomnia  \cite{espie_kyle_williams_ong_douglas_hames_brown_2012,Freeman_Waite_Startup_Myers_Lister_McLnerney_2015, luik_digital_2017}, in which users are invited to submit daily sleep diaries that help inform the treatment users receive. The widespread use of Sleepio means large quantities of records are being generated daily, which creates an unique opportunity to train neural networks and model the relationship between user behaviour and sleep quality \cite{lecun_deep_2015, Goodfellow_et_al_2016}. 

Neural networks are the model of choice to produce user recommendations when large amounts of records tracking user behaviour are available \cite{covington_deep_2016, ying_graph_2018}, but to our knowledge such an application is lacking in the use of digital sleep diaries. Besides the need for large quantities of data, other reason why neural networks have not yet been applied in this context is that clinicians may not use a recommendation given by an opaque model such as a neural network unless they know (i) the confidence that the model has on each particular prediction, and (ii) the reasons priming the model to give each particular recommendation \cite{obermeyer_predicting_2016, chen_machine_2017}.

In this study we train a neural network on Sleepio diaries from a random sample of 20,000 users to predict sleep quality from sleep-wake behaviors (e.g., alcohol consumption, exercise, caffeine, noise and light exposure). We then use the gradient of the output of the neural network (i.e. estimated sleep quality) with respect to its input (i.e. reported sleep behaviour during the last 10 days) to build recommendation systems that give personalised advice to each user. We show that users who follow the advice given by the neural network report significantly better than average sleep quality, and better than those users that follow recommendations given by linear models and other algorithms. We further build modules in the neural network that express the confidence held on each prediction by outputting confidence intervals, and we then show that these intervals are accurately capturing the distribution of the population of users around each prediction. Finally, we demonstrate that with first and second order derivatives we can explain what factors of sleep behaviour are priming the neural network to predict bad or good sleep quality for each user, and that these explanations can be calculated across the average of the population and for each individual user.

\section{Method}

\subsection{Data}
Data consists of 401,174 self-reports of sleep behaviour and sleep quality ratings on a condensed digital version of the Consensus Sleep Diary \cite{carney_buysse_ancoli-israel_edinger_krystal_lichstein_morin_2012} generated by 20,000 random users of Sleepio \cite{cowie_multimedia_2018, elison_feasibility_2017, carney_buysse_ancoli-israel_edinger_krystal_lichstein_morin_2012}. The mean number of daily reports per user is 20.1, while percentiles 10/50/90 are 3/9/34 reports per user. Each report includes 12 numeric and 11 binary variables describing sleep behaviour (see table \ref{descriptives}), while sleep quality is scored by a 5 level variable that ranges from -2 (worst sleep) to +2 (best sleep).

\begin{table}[h!]
  \renewcommand{\arraystretch}{1.2}
  \caption{\textbf{Descriptive statistics on used dataset.} Table lists standard statistics on each of the Digital Sleep Diary variables and additional behavioural variables present in the dataset we used from Sleepio. For numeric variables, mean, percentile 10 (P10), and 90 (P90) are shown. For binary variables, count and proportion (\%). }
  \label{descriptives}
  \centering\scriptsize
  \begin{tabular}{c||c||c}
  \small
  Variable   & Type   & Mean [P10, P90] / count [\%] \\
  \hline
    date    & day    & 15.7 [ 3, 28 ]  \\
    date     & month    & 5.9 [ 2, 11 ]  \\
    date     & year    & 2017.1 [ 2016, 2018 ]  \\
    time into bed     & hour    & 22.4 [ 18, 3 ]  \\
    time into bed     & mins    & 17.4 [ 0, 45 ]  \\
    time into bed before lights out     & mins    & 24.1 [ 0, 60 ]  \\
    sleep onset latency     & mins    & 27.6 [ 0, 60 ]  \\
    time awake at night     & mins    & 30.5 [ 0, 90 ]  \\
    total sleep time     & mins    & 418.1 [ 285, 540 ]  \\
    total time in bed     & mins    & 31.0 [ 0, 75 ]  \\
    times awake at night     & count    & 2.0 [ 0, 5 ]  \\
    no sleep obtained     & binary    & 377 [ 0.6 \% ]  \\
    subjective sleep quality     & 5-levels score    & -0.2 [ -1, 2 ]  \\
    note written     & binary    & 10515 [ 18.1 \% ]  \\
    use of alcohol      & binary    & 2678 [ 4.6 \% ]  \\
    use of caffeine      & binary    & 799 [ 1.4 \% ]  \\
    exercise      & binary    & 1163 [ 2.0 \% ]  \\
    lights on      & binary    & 680 [ 1.2 \% ]  \\
    use of nicotine      & binary    & 218 [ 0.4 \% ]  \\
    noise      & binary    & 1400 [ 2.4 \% ]  \\
    pain      & binary    & 1687 [ 2.9 \% ]  \\
    slept with partner      & binary    & 1794 [ 3.1 \% ]  \\
    sleeping pills      & binary    & 2463 [ 4.2 \% ]  \\
    temperature      & binary    & 1939 [ 3.3 \% ]  \\
  \hline
  \end{tabular}
 \end{table}

\subsection{Linear Model of Sleep Quality}

With systematic backwards elimination \cite{zhang_variable_2016} we search for the linear model that best estimates the sleep quality of the last recorded day in Sleepio as a function of the following: the value of all other variables on that last day (except sleep quality itself), the value of all variables on the previous day (including sleep quality), the average value of all variables in the 10 day window preceding the last day (excluding from this average the sleep quality of the last day). Backward elimination was stopped once the Akaike Information Criterion (AIC) of the model ceased to decrease \cite{lovric_akaikes_2011, wagenmakers_aic_2004}. The optimal linear model found is described in supplementary materials, section 1.

\subsection{Neural Network Model of Sleep Quality}

We built a neural network to model sleep quality (output variable ‘qualityPrediction’ in figure \ref{architecture}) as a function of the sleep behaviour of users during the previous 10 days (input variable ‘behaviour’), in addition defining a flag per sleep behaviour input with missing values (input variable ‘miss’). The first layer of the architecture (‘zscore’ in figure \ref{architecture}) centres and standarises each variable across users and time. The second layer (‘missing\_mask’) creates an attention mask from the missing value flags, and then applies this mask to the output of the first layer. A third layer (‘LSTMs’) consists of 2 stacked LSTMs with tanh activation function and 50 and 10 neurons, respectively. Fourth layer (‘dense’) applies to the last output of the stacked LSTMs a dense transformation with activation function elu, producing a single output per user. Finally, the fifth layer (‘rescale\_quality’) rescales the output for it to match the scale [ -2, +2 ] of sleep qualities used by Sleepio. In addition to predicting sleep quality itself, with modules ‘dense\_intervals’ and ‘rescale\_intervals’ the neural network also estimates 9 confidence intervals of sleep quality, which correspond to the 0.1, 0.2 ... 0.9 probabilities of the actual sleep quality of each user laying within the estimated 1st, 2nd, ... 9th interval. 

\begin{figure*}
  \label{architecture}
  \centering
  \includegraphics[ scale = 0.7 ]{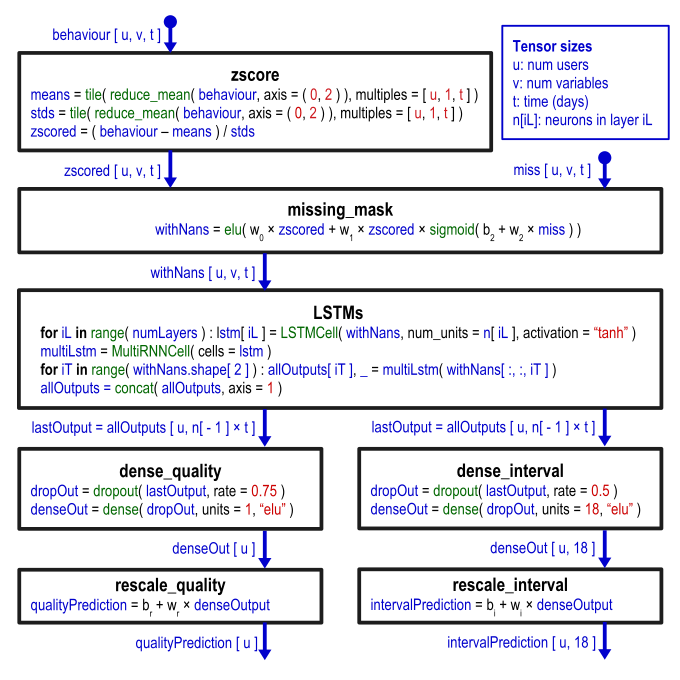}
  \caption{ \textbf{Architecture of neural network.}  Each module of the neural network is represented by a rectangle, with the name of the module in black bold letters, its corresponding tensorflow code below, function names in green, variable names in blue, and constants in red. Tensors exchanged by operations are represented by blue arrows, with the size of each tensor in brackets. A legend in the upper right corner lists the meaning of letters used to represent tensor sizes. }
  \label{architecture}
\end{figure*}

The neural network is trained and tested with 10-fold cross-validation on the 5,809 users, out of a total of 20,000, that did report sleep quality on their last day on record. The training algorithm is Adam with a learning rate of 0.01, and batch size 256. The loss function is the sum of two components. The first component is the mean squared error on predicting sleep quality:

\begin{equation}
\sum \limits_{u\in users} \big( y_{n}(u) - y_{s}(u) \big) ^2
\end{equation}

where $y_n(u)$ is the sleep quality predicted by the neural network for user  (output 'qualityPrediction' in figure \ref{architecture}), while $y_s(u)$ is the sleep quality reported by the user in Sleepio application. The second component of the loss function is mean error on the correct proportion of subjects falling within each confidence interval, whose boundaries $b_{low}(...)$ and $b_{high}(...)$ are predicted by the neural network (output 'intervalPrediction' in figure \ref{architecture}):

\begin{equation}
    \begin{split}
        \sum \limits_{i=1...9} \bigg[ \sum \limits_{u\in users} &  sig_{10}( b_{low}(i) -y_{n}(u)) + sig_{10}( y_{n}(u)- b_{high}(i) ) \bigg] - p(i) 
    \end{split}
\end{equation}
where $i$ indexes the 9 confidence intervals (which respectively correspond to the 0.1, 0.2... 0.9 probability confidence intervals), and $u$ indexes the users of Sleepio. The function $sig_{10}$ is simply the sigmoid function with its input multiplied by 10:

\begin{equation}
sig_{10}(x) = \frac{ 1 }{ 1 + e^{-10x} }
\end{equation}

This sigmoid function transforms the process of counting how many samples are within $b_{low}(...)$ and $b_{high}(...)$ into a soft function with well defined derivatives

\subsection{Neural network recommendations of sleep behaviours from sleep diaries}

We create and test one recommender of sleep behaviour for each one of our 2 models of sleep quality (the neural network, and the linear model). Given a model that estimates sleep quality ($y$) as a function of other sleep diary variables ($\vec{x}$), in order to find which values of $\vec{x}$ produce the highest $y$, we follow the gradient $dx/dy$ in the direction that increases $y$. While the gradient $dx/dy$ depends on all the input variables $\vec{x}$, when following the gradient we only change the values of $\vec{x}$ that correspond to the last day recorded ($\vec{x}(t=0)$). In this manner we find the values that, according to the model, the input variables of the last day should have for the output $y$ (sleep quality) to be maximal. In the case of the linear model, where all gradients $dy/dx$ are constant, this process (i.e. following the gradient) would continue without bound and in the same direction in the ${\rm I\!R}^{n}$ space defined by $\vec{x}$. To prevent this, we stop following the gradient once the model estimates that the current values of $\vec{x}$ already produce sleep quality +2, which is the maximum of the scale and corresponds to 'best sleep'. 

In addition to the two recommenders (one from the linear model, another from the neural network) derived from a model with the process described above, we also create two other recommenders. One of these two recommenders, which we call ‘best neighbourhood’, attempts to find the behaviour that is most helpful in average across the population without the use of any model. This first recommender, called 'best neighbourhood',  z-scores all behaviour variables across the population, then selects 1000 random users, for each of these users finds the 100 neighbour users that are closest in sleep behaviour of the last day (euclidean distance), and then calculates the average sleep quality of those 100 users - we call this average the ‘neighbourhood sleep quality’. Among the 1000 initially selected random users, the one with the highest ‘neighbourhood sleep quality’ is chosen as ‘best sleeper’, and its behaviour on the last day on record will be the recommendation given by the recommender to everybody else in the population. The second recommender, called ‘best day’, recommends to each user to follow the behaviour that the user had on the day when he/she slept best on record.

Each recommendation given by the commender systems consists on a value for each of the following variables: ‘caffeine’, ‘noise’, ‘nicotine’, ‘lights on’, ‘slept with partner’, ‘alcohol’, ‘time in bed with lights off to asleep’, and ’time in bed with lights on’. To evaluate each recommender system, we calculate the average sleep quality of the users as a function of how many of the recommendations they ignored. For instance, given the recommendation [ ‘do not drink alcohol’, ‘do not take caffeine’, ‘have exercise’ ], a user that in the last day on record did [ ‘drank alcohol’, ‘did not take caffeine’, ‘had exercise’ ] would have ignored 1 recommendation, while a user that did [ ‘drank alcohol’, ‘took caffeine’, ‘had exercise’ ] would have ignored 2. However, these three examples only consider binary variables, while recommendations are also made on two numeric variables: ‘time in bed with lights off to asleep’ and ’time in bed with lights on’. For these, we count how many of the behaviours of the user were >30 minutes away from the recommendation.


\section{ Results }

\subsection{Recommendations of sleep behaviour}

As described in methods, we create one recommender of sleep behaviour from each of the models of sleep quality by following the gradient $dy/dx$. The recommender systems that we evaluated were based on: a linear model, a neural network, the best neighbourhood average quality, and the best day quality. Each recommender advises what sleep behaviour each user should follow on their last day on record for the sleep behaviour variables: ‘caffeine’, ‘noise’, ‘nicotine’, ‘lights on’, ‘bed partner’, ‘alcohol’, ‘time in bed with lights off to asleep’, and ’time in bed with lights on’

\begin{figure*}
  \centering
  \includegraphics[width=0.95\textwidth]{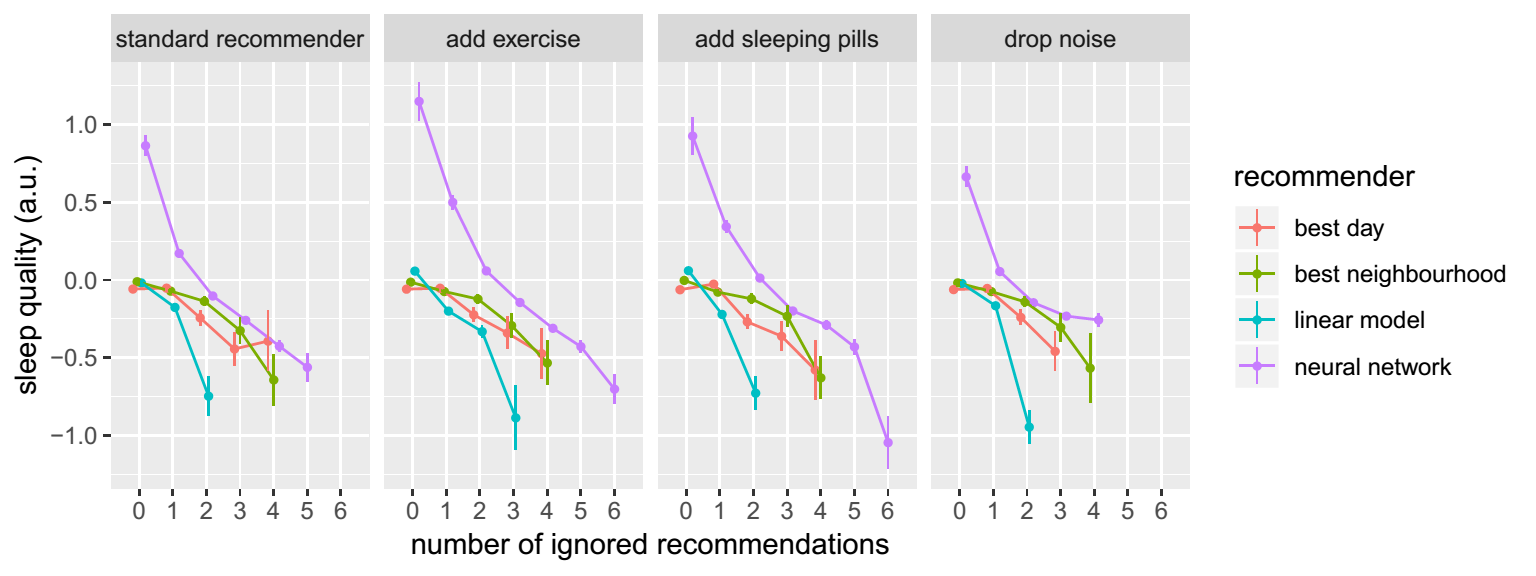}
  \caption{ \textbf{Effectiveness of recommenders.}  The figure shows average sleep quality of users (y axis) as a function of how many recommendations they ignore (x axis) for each recommender (colour). Points represent average across users, while vertical bars represent standard error. Leftmost panel shows effectivity for standard recommender, while other panels shows effectivity after introducing 3 modifications into the standard recommender. Namely, from left to right: allowing the recommender to also advise on exercise; allowing also advice on taking sleeping pills; and barring advice on background noise. }
  \label{compare}
\end{figure*}

To estimate how effective the recommenders are, we calculate the average sleep quality of users as a function of how many of the recommendations they ignored (see figure \ref{compare}). Users that followed all recommendations from the neural network reported an average sleep quality of $0.86\pm0.065$ (mean $\pm$ standard error), which was significantly better than the sleep quality of users following the linear model (p-value $3.5\cdot10^{-42}$), the best neighbourhood ($6.5\cdot10^{-40}$), or the best day recommenders ($2.1\cdot10^{-44}$). The question remains whether it is fair to recommend users regarding noise, as in occasions users might not have control over noise level in their environment. We therefore also estimated the effectiveness of the recommender when it did not give recommendations regarding noise, and observed again better sleep quality on users following the neural network ($0.66\pm0.068$) than on users following any of the other systems. Another question is whether to include the use of sleeping pills (in the preceding night) in this model as this recommendation may not be appropriate from an ethical and clinical standpoint. However, if given this option, a neural network also recommending on sleeping pills gave an average sleep quality of $0.92\pm0.096$, again better than all other recommenders under the same circumstances. Finally, if we allowed recommenders to also advise on the behaviour variable ‘exercise’, users that followed all recommendations of the neural network reported average sleep quality of $1.25\pm0.11$, which was better than all other systems.

\subsection{Explaining the neural network model}

\begin{figure*} [h!]
  \centering
  \large{A}
  \includegraphics[width=0.4\textwidth]{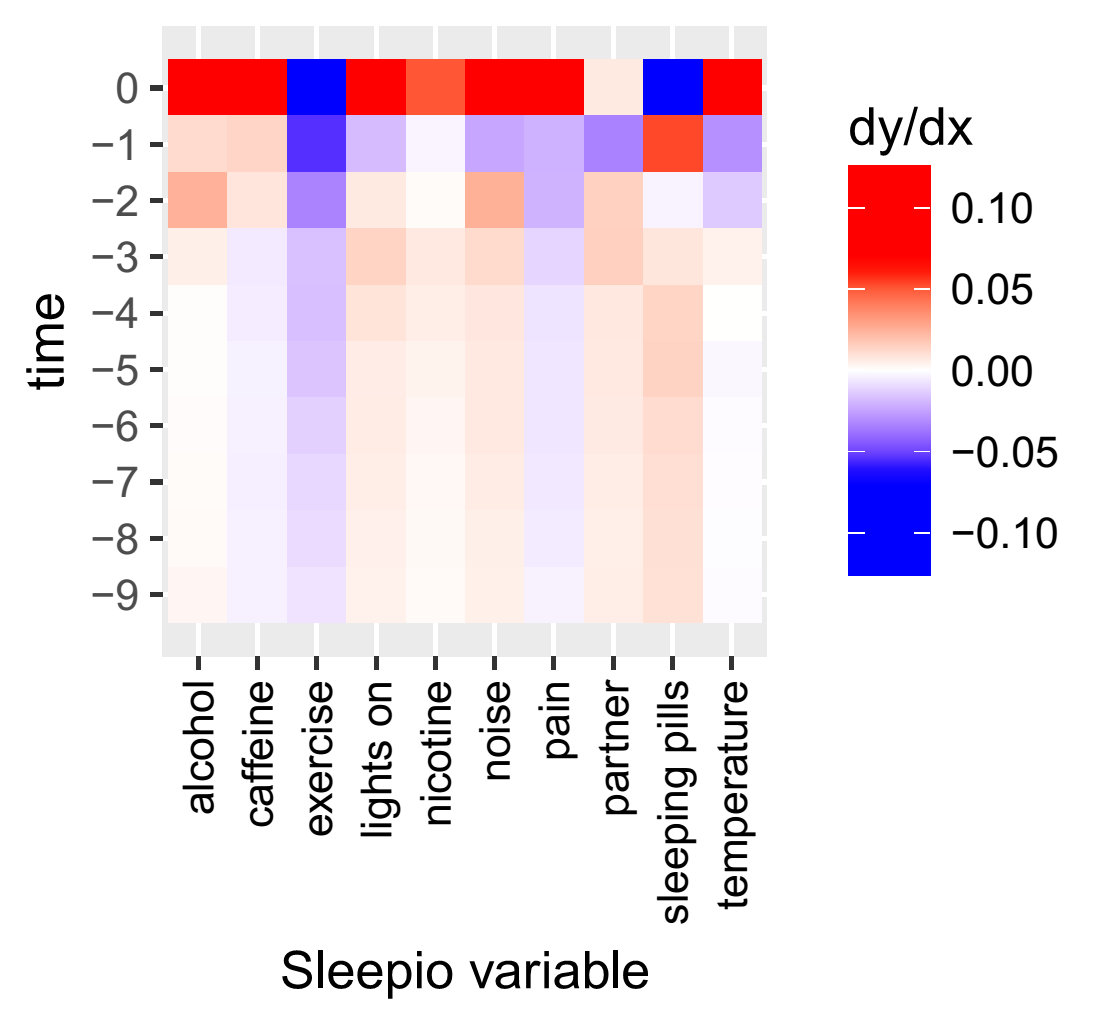}
  \vspace{5 mm}
  \large{B}
  \includegraphics[width=0.25\textwidth]{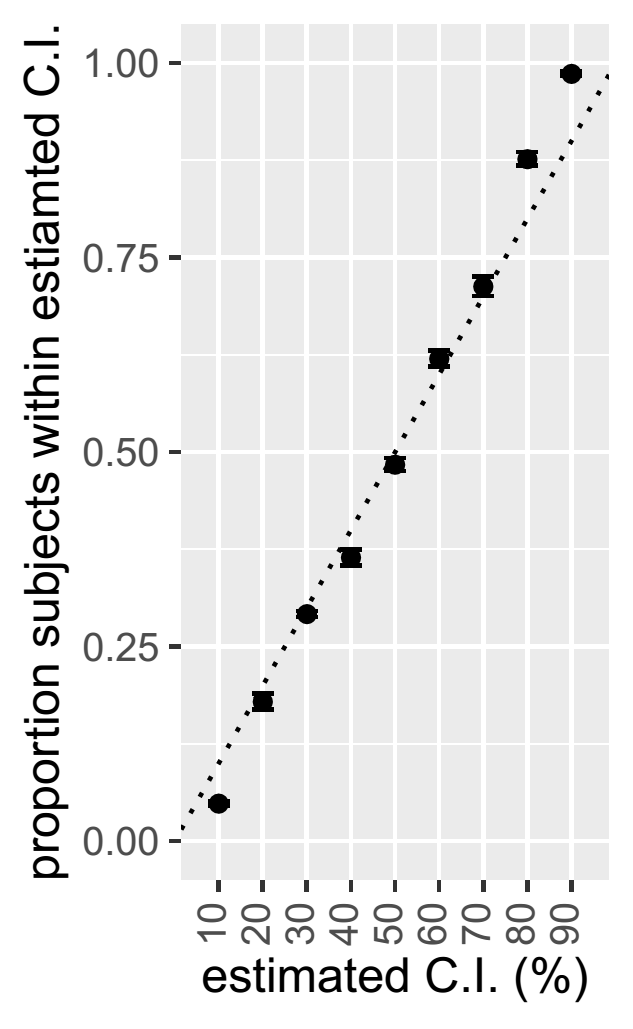}
  \caption{ \textbf{Confidence and explainability of the neural network.}  A:  Colour represents average first order derivatives of the sleep quality predicted by the neural network (output ‘qualityPrediction’ in figure \ref{architecture}) with respect to each of the binary sleep behaviour variables (input ‘behaviour’ in figure \ref{architecture}). The average on the derivative is calculated across users, as for each user these derivatives change depending on other user behaviours imputed to the neural network. X axis lists each of the behaviour variables, while y axis lists of the time steps that the neural network reads. B: Proportions of users (y axis) whose last day sleep quality falls within each of the confidence intervals calculated by the neural network (x axis, output intervalPrediction in figure \ref{architecture}). The diagonal that would correspond to a perfect prediction is shown with dashed lines. }
  \label{intervals}
  \label{der1}
\end{figure*}

As noted in the introduction, model explainability is a central requirement made by clinicians. To explain how the neural network estimates sleep quality as a function of sleep behaviour, we calculate the first derivative of the output as a function of each of its inputs ($dy/dx$), where the output ‘y’ is the estimated sleep quality and the inputs ‘$\vec{x}$’ is sleep behaviour. According to these derivatives, the strongest influence on sleep quality comes from the sleep behaviour of the last day (see figure \ref{der1}A). The influence of most variables changes direction from last to previous to the last day. For instance, the use of sleeping pills on the preceding night has a positive influence on sleep quality ($dy/dx$ = -0.090), while use of sleeping pills of the previous to last nights has a negative impact ($dy/dx$ = 0.053). Other variables do not change direction of influence across days. For instance, exercising consistently has a positive impact on sleep quality independently of how many days ago it took place (e.g. $dy/dx$ = -0.075, -0.054, -0.032 for the last 3 days).

To further explain the behaviour of the neural network, second order derivatives ($dy^2/(dx\_1 dx/2)$) were also calculated. Their values were in general of lower magnitude than the first order derivatives, but still significant, specially if $x_1$ and $x_2$ are sampled on the same day (see supplementary material, section 2).

Finally, the second requirement made by clinicians is to measure how confident the system is on each recommendation. With this objective, the neural network also calculates 9 confidence intervals (CI) per recommendation. Each of the 9 CIs consists on a window within which a given percentage of the time the actual sleep quality will fall. For instance, if for the 20\% CI the neural network outputs 0.2 and 0.5, it means that the neural network estimates that it is 20\% likely that the actual sleep quality will be between 0.2 and 0.5 if the user follows that recommendation. We observe that the proportion of sleep qualities falling within each interval correlates strongly with the probabilities that each interval tries to capture (correlation 0.99, p-value $1.32\cdot10^{-9}$, see figure \ref{intervals}B), and they can be used as a direct proxy of model confidence.

\begin{figure*}
  \centering
  \large{A}
  \includegraphics[width=0.23\textwidth]{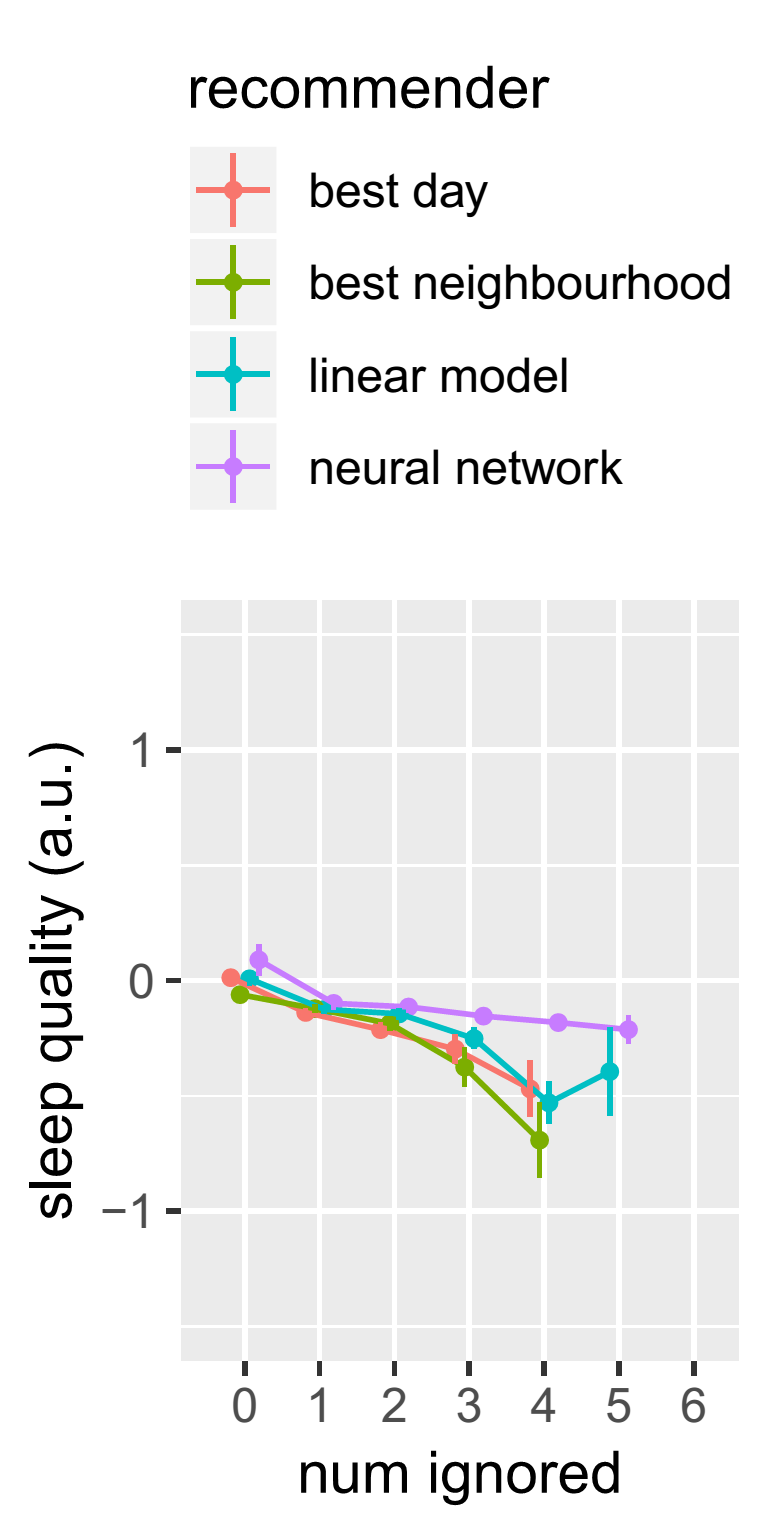}
  \large{B}
  \includegraphics[width=0.23\textwidth]{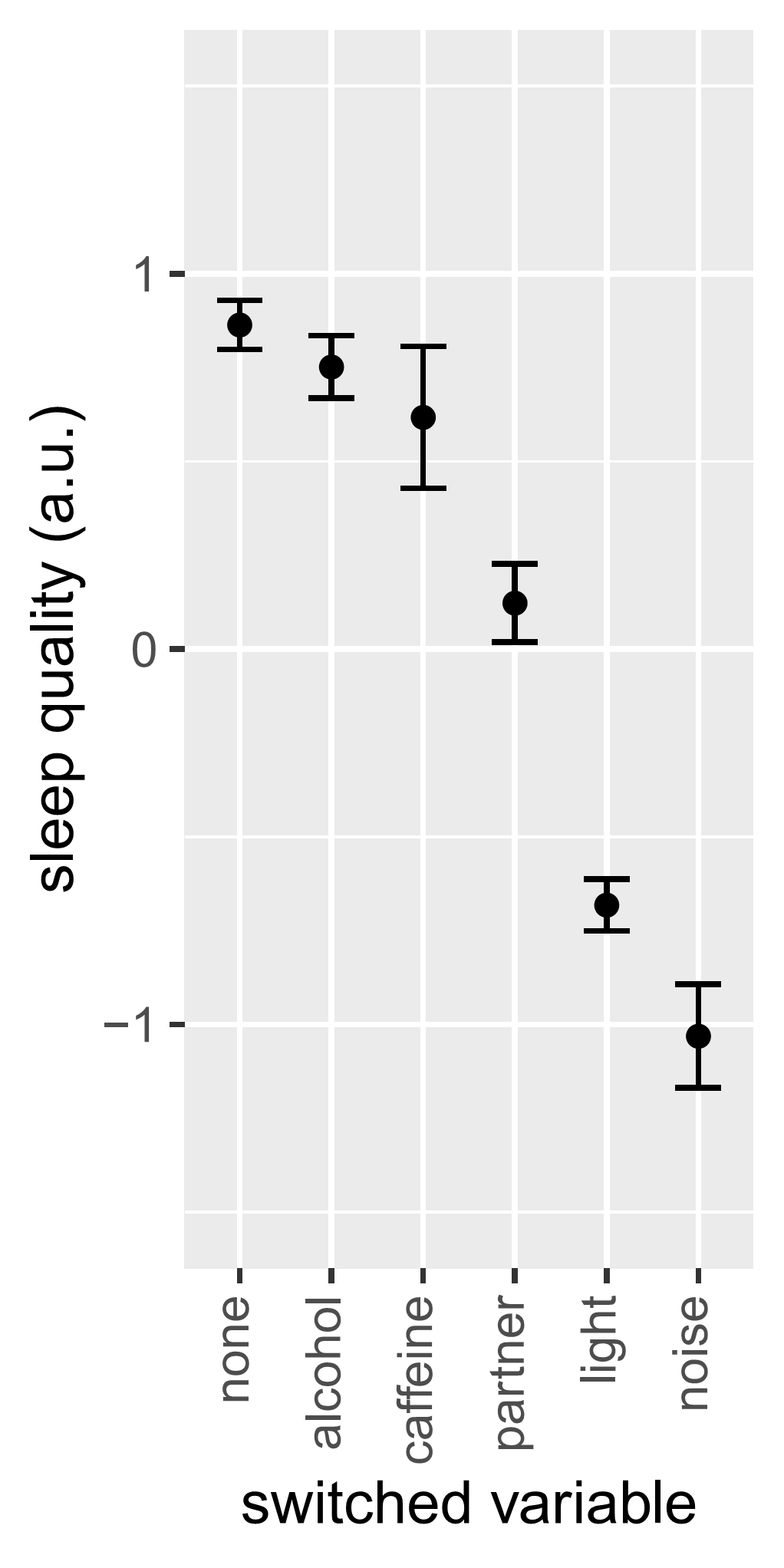}
  \large{C}
  \includegraphics[width=0.23\textwidth]{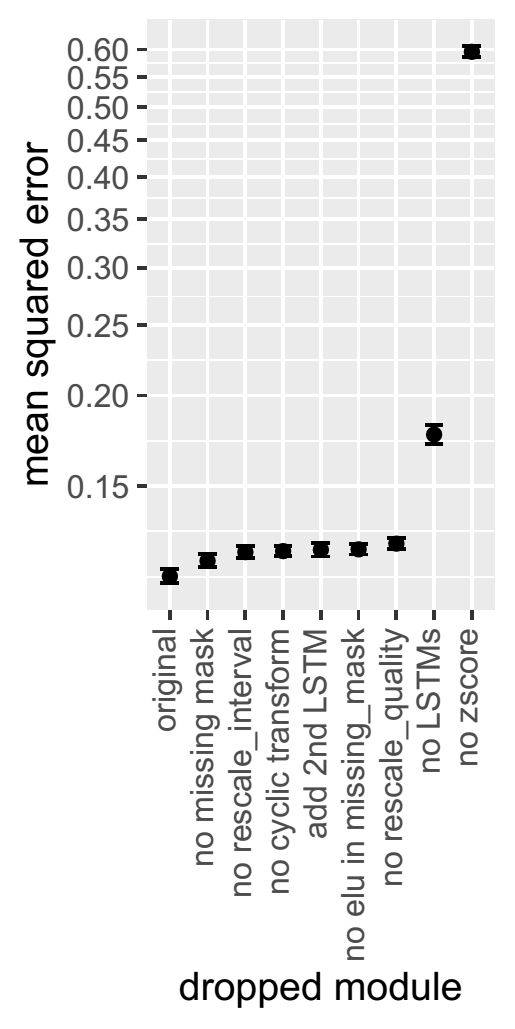}
  \caption{ \textbf{Robustness tests on the neural network.}  A: Same neural network and diagram as in figure \ref{compare}, standard recommender, but after randomly allocating advice to users. B: Average sleep quality (y axis) of users that follow all recommendations of the neural network after switching one of each of the binary variables (x axis). For all figures, points represent averages, and vertical bars standard errors. C: Mean squared error (y axis) of estimating sleep quality for a number of modifications on the neural network (x axis). }
  \label{shuffle}
  \label{flip}
  \label{ablation}
\end{figure*}

\subsection{ Robustness of neural network }

A question remains whether the neural network is actually personalising its recommendations to the historic behaviour of each user, or whether it is simply giving better global recommendations than those calculated by the other recommenders. To test this, we shuffled the recommendations that the classifiers built for each user, and then we re-calculated the average last day sleep quality of users as a function of number of ignored recommendations. Namely, while in previous sections we used for each user the advice that the recommender designed for him/herself, here we used for each user the advice that the recommender designed for another randomly selected user. We observe that, after shuffling, the sleep quality of users following the neural network drops to $0.01\pm0.052$ (see figure \ref{shuffle}A), which is not distinguishable from the linear model (p-value 0.55) or best neighbourhood (0.71) with this number of samples. While shuffling users significantly decreased the quality of the neural networks ($1.5\cdot10^{-41}$), a change was not observed for linear model (0.23) or best neighbourhood (0.74).

Another question is whether the neural network is optimal on the recommendations given to each individual behaviour variable. Namely, the recommender may not lie on a local optimum, and further tuning of individual variables may further improve recommendations. To test this possibility we re-calculate average sleep quality of users that followed all the recommendations of the neural network, but after switching the recommendation given on each individual variable, one at a time. We do this only with binary variables, as it is not trivial how to ‘switch’ recommendations on a non-binary variable. Compared to the original recommendations (‘none’ in figure \ref{flip}B), switching alcohol (0.074) or caffeine (0.15) produces a not statistically significant decrease in sleep quality, while sleeping with bed partner ($6.4\cdot10^{-10}$), lights on ($4.6 \cdot10^{-41}$) or noise ($1.8\cdot10^{-14}$) produces a significant decrease.

Finally, we also measured the contribution that different modules of the neural network had to its accuracy to predict sleep quality (0.11 mse of equation 1 for the original architecture, see figure \ref{ablation}C). We observed that eliminating the mask of missing values in layer ‘missing\_mask’ has no significant effect on mean squared error of sleep quality (p-value 0.13). No significant difference was also found if we did not transform cyclic input variables via sine and cosine functions (0.082). However, performance decreased if eliminating module 'rescale\_interval' (0.0034), if we did not transform cyclic input variables via sine and cosine functions (0.0021), adding a second layer of 10 units to the LSTMs module (0.0019), eliminating elu transform on module ‘missing\_mask’ (0.0018), eliminating module 'rescale\_quality' (0.0010), removing all LSTMs ($2\cdot10^{-3}$), or removing module ‘zscore’ ($4\cdot10^{-5}$). Besides architecture, the design of the loss function may also have an effect on prediction performance. In particular, the fact that we are trying to optimise predictions on sleep quality and confidence intervals may have a detrimental impact on the loss achieved by the neural network. However, we actually find that mse for predicting sleep quality (equation 1) significantly worsens (mse $0.64\pm 0.03$) if we attempt minimising only on sleep quality loss (equation 1) rather than minimising on the full loss (the sum of equations 1 and 2)


\section{Declaration of Competing Interest}
Authors Colin Espie, Thomas Walker, Chris Miller, and Alasdair Henry have contracts with Bighealth ltd. All other authors state no conflicts of interest.

\section{Acknowledgements}
Research reported in this publication was supported, in part, by the National Institute for Health Research (NIHR) Artificial Intelligence for Health and Social Care Award (AI-AWARD02183), the Virtual Brain Cloud award (H2020-SC1-DTH-2018-1; Grant agreement ID: 826421), and the EPSRC Center for Doctoral Training in Health Data Science (EP/S02428X/1).

\bibliographystyle{unsrtnat}
\bibliography{literature}

\section{Appendix}

\subsection{ Linear model of sleep quality }

\begin{figure*}[!h]
  \centering
  \large{A}
  \includegraphics[width=0.7\textwidth]{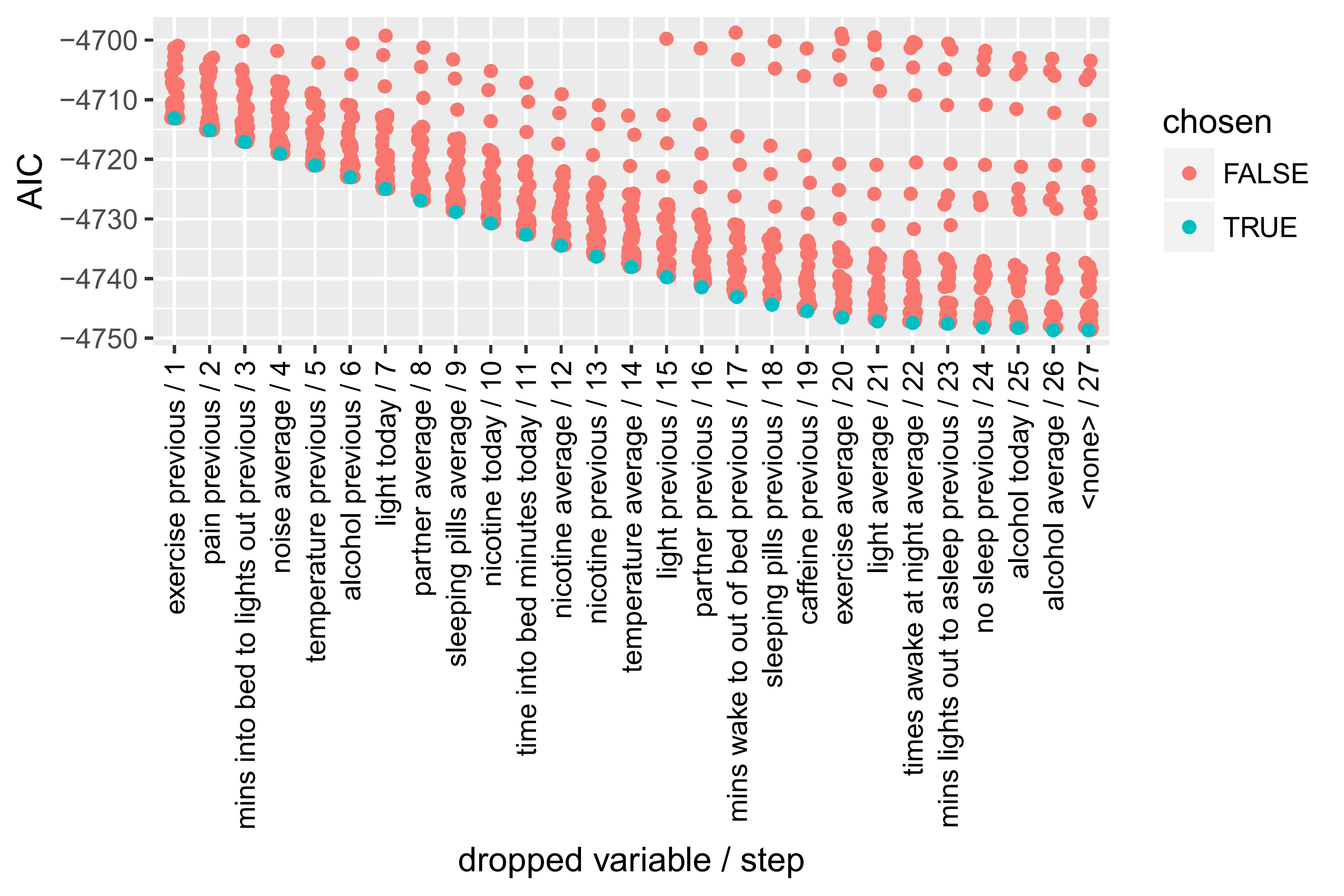}
  \vspace{5mm}

  \large{B}
  \includegraphics[width=0.7\textwidth]{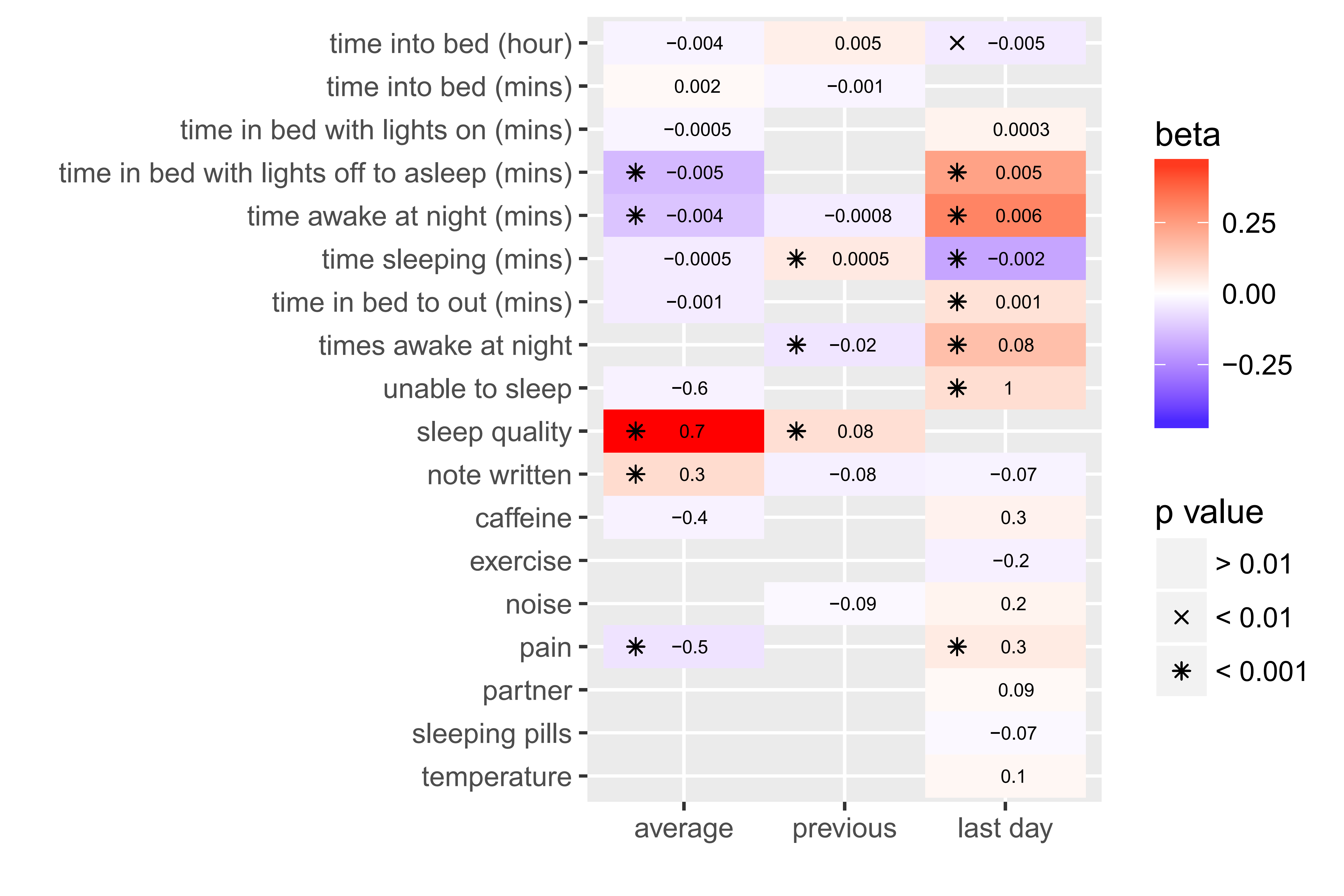}
  \caption{ \textbf{Stepwise selection and linear model.} A: A backward stepwise selection algorithm was allowed to attempt at modelling “sleep quality” with any combination of variables (rows) from last day on record (3rd column), previous to last day on record (2nd column) or their global average value (1st column) per patient. The selection algorithm calculated AIC (y-axis) on the model using all these variables and on the same model after dropping each variable one at a time. The algorithm will then select this model that had lowest AIC (green points), discard all others (red points) and restart dropping each remaining variable and recalculating AIC. This process was repeated (x-axis) until no further improvement on AIC was observed. B: Fitting results of the final model built by stepwise selection. Beta-values (cell colour) and b-values (written in each cell - this is equal to beta-value but scaled to the units of each variable) are shown for each behaviour variable (rows) from last day on record (3rd column), previous to last day on record (2nd column) or their global average value (1st column) per patient. The statistical significance of each input is represented by a mark - ‘*', 'x' or ‘ ‘ depending on p-value. For variables not chosen by stepwise selection, their corresponding cell is left blank. }
  \label{lm}
\end{figure*}

In order to find the linear model that best estimates sleep quality of the last day on record, we started by using as inputs the values of all sleep diary  variables from the last day, from the day previous to the last day, and from the average of the last 10 days. Systematic backwards elimination then sequentially dropped variables until it found no further improvement of AIC after eliminating 26 variables, at which point AIC was -4748 and R2 0.58. With this model, 16 out of 36 variables were significant at p-value threshold 0.01 (multiple comparisons correction with bonferroni), and 15 at p-value 0.001 (see figure \ref{lm}).

\subsection{ Second order derivatives of the neural network }

A five layer neural network estimates sleep quality of the last day on record by using as input all sleep dairy variables from the last 10 days (tensor ‘behaviour’ in architecture figure). The neural network also estimates 9 confidence intervals, which correspond with the 0.1, 0.2, … 0.9 probabilities of the actual sleep quality reported by each user falling within such interval. 

\begin{figure*}[t!]
  \centering
  \large{A}
  \includegraphics[width=0.45\textwidth]{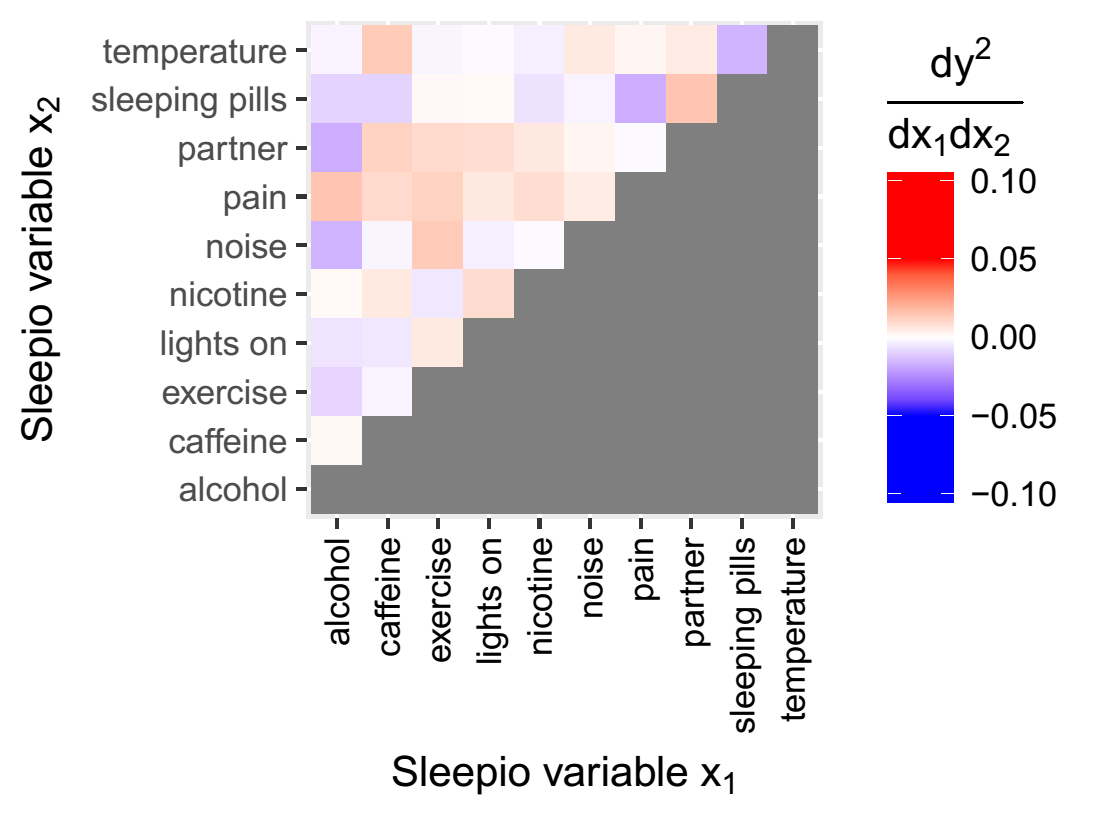}
  \large{B}
  \includegraphics[width=0.45\textwidth]{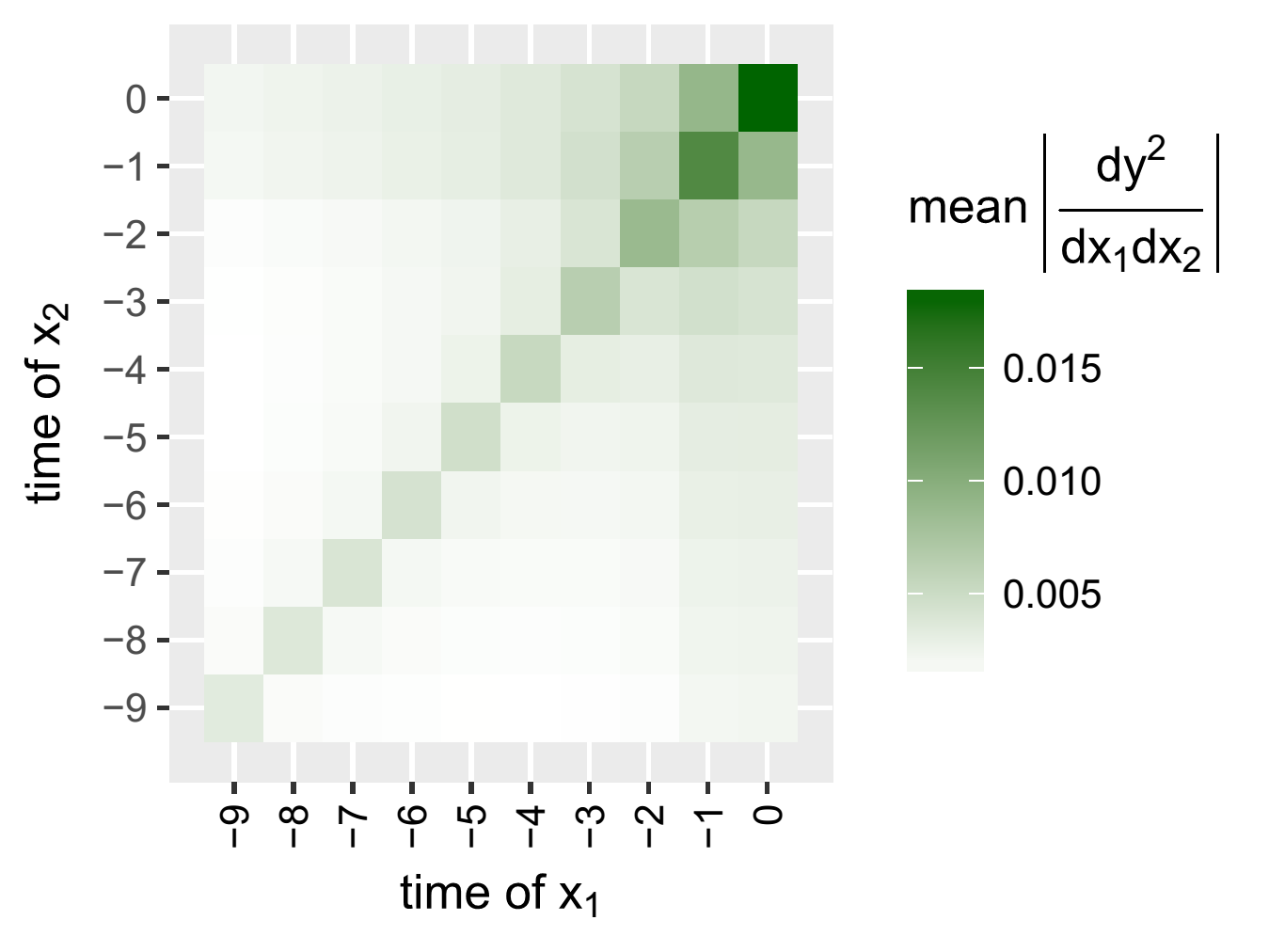}
  \caption{ \textbf{Second order derivatives of the neural network.} A: Average second order derivatives (colour) within time step 0 (i.e. last day in record for each user) - namely, $dy_2/(dx_1(t_1=0)dx_2(t_2=0))$. The average is calculated across users. B: Average second order derivative for each combination of time steps. Averages are calculated across users and across all sleepio variables ($x_1$ and $x_2$), and on the absolute value of derivatives rather than signed derivatives, as otherwise positive and negative derivatives would cancel each other in the average. }
  \label{der2}
\end{figure*}

On explaining how the neural network estimates sleep quality (mathematically, building the function $f(\vec{x})=y )$, the first order derivatives only explain the linear component of how the neural network processes $\vec{x}$ to approximate y (i.e. the linear component of $f(\vec{x})=y$, or first order Taylor expansion). However, the expression capacity of neural network is well beyond linear functions, and they are expected to make substantial use of interactions and nonlinearities to build a more accurate estimation of sleep quality. After first derivatives, the next largest contribution to explain how the neural network is approximating sleep quality ($y$) are the second order derivatives ($dy^2/dx\_1 dx/2$). There is no mathematical equivalent of second order derivatives in linear model. When calculated for our neural network, second order derivatives have on average a weaker effect on the estimation of sleep quality ($y$) (see figure \ref{der2}) than the effect of the first order derivatives that were closest to last day on record $(dy/dx(t>-2)$). The effect that these second order derivatives have on estimated quality are, by definition, additive to the first order derivatives, and can be understood as finer corrections to the first order ones. When averaged across sleep behaviours (i.e. across $x\_1$ and $x\_2$), it becomes clear that their additive effects are maximal when x\_1 and x\_2 are sampled on the same day (see figure \ref{der2}). The interpretation is that, according to the neural network, sleep behaviours tend to interact within the same day, but not across days. Namely, the effect that taking alcohol today and having taken sleeping pills 2 days ago have on sleep quality is independent of each other. However, the effect of taking alcohol today is statistically dependent on the effect of taking sleeping pills also today, and they interact on their impact on sleep quality.

\end{document}